\pdfoutput=1

\documentclass[11pt]{article}

\usepackage[preprint]{acl}

\usepackage{times}
\usepackage{latexsym}
\usepackage{booktabs}
\usepackage{amsmath}
\usepackage{subcaption} 
\usepackage{enumitem}
\setitemize{noitemsep,topsep=0pt,parsep=0pt,partopsep=0pt}

\usepackage[T1]{fontenc}

\usepackage[utf8]{inputenc}

\usepackage{microtype}

\usepackage{inconsolata}

\usepackage{graphicx}
\usepackage{amsfonts}

\title{Fast Forwarding Low-Rank Training}

\author{Adir Rahamim \\
  Technion – Israel Institute of Technology \hspace{5mm} \\
  \texttt{adir.rahamim@campus.technion.ac.il} \\ \And
  Naomi Saphra \\
  \hspace{5mm} Kempner Institute at Harvard University \\
  \texttt{nsaphra@fas.harvard.edu} \AND
  Sara Kangaslahti\\
  Harvard University\\
  \texttt{sarakangaslahti@g.harvard.edu}
 \And
 Yonatan Belinkov\\
  Technion – Israel Institute of Technology \hspace{5mm} \\
  \texttt{belinkov@technion.ac.il}
  }

\begin{document}
\maketitle
\begin{abstract}

Parameter efficient finetuning methods like low-rank adaptation (LoRA) aim to reduce the computational costs of finetuning pretrained Language Models (LMs). Enabled by these low-rank settings, we propose an even more efficient optimization strategy: Fast Forward, a simple and effective approach to accelerate large segments of training. In a Fast Forward stage, we repeat the most recent optimizer step until the loss stops improving on a tiny validation set. By alternating between regular optimization steps and Fast Forward stages, Fast Forward provides up to an 87\% reduction in FLOPs and up to an 81\% reduction in train time over standard SGD with Adam.
We validate Fast Forward by finetuning various models on different tasks and demonstrate that it speeds up training without compromising model performance. Additionally, we analyze when and how to apply Fast Forward. 

\end{abstract}

\section{Introduction}

Modern optimizers provide a spectacular array of tweaks to stabilize training trajectories and accelerate Stochastic Gradient Descent (SGD). 
Yet even with every optimization hack in the modern machine learning toolkit, the expense of training accumulates. In this paper, we ask: \textit{what if we just keep going in the same direction until it stops helping?}

\begin{figure}[t]
    \centering
    \includegraphics[width=.65\linewidth]{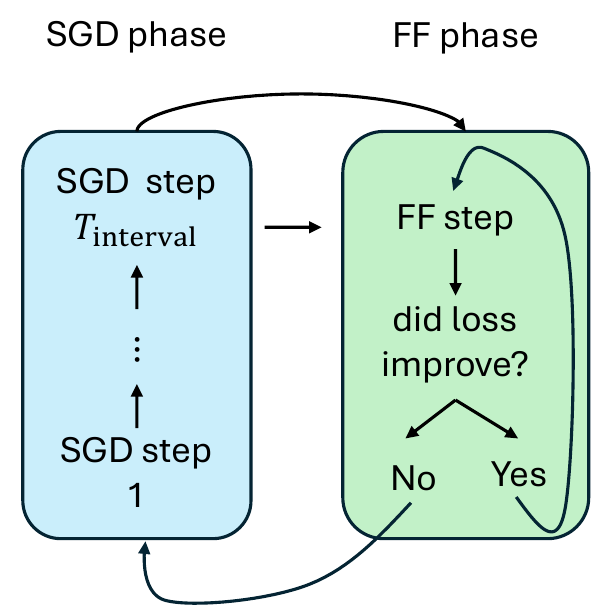}
    \caption{Fast Forward algorithm. We alternate between SGD and Fast Forward, exiting the Fast Forward stage when the loss on a tiny validation set stops improving. }
    \label{fig:method}
\end{figure}

Applying this exceedingly simple approach, which we call Fast Forward, to low-rank training (Section \ref{sec:background}) allows enormous speedups over standard optimization.  We alternate between Adam SGD for burn-in and accelerating by line search with a tiny validation set, which provides an ad-hoc optimal step size much larger than that determined by the learning rate. (See Figure \ref{fig:method} and Section \ref{sec:fast_forward}.) 


We experiment with Fast Forward on three finetuning tasks and four language models ranging from 1.4 to 6.9 billion parameters (Section \ref{sec:experiments}). In all cases, we find consistent efficiency gains with Fast Forward (Section \ref{sec:results}), reaching the performance of regular low-rank finetuning \textbf{41--87\% faster}.

While Fast Forward works incredibly well in low-rank finetuning, it fails to improve full-rank standard fine-tuning (Section \ref{sec:results_fullrank}). We investigate possible causes, showing that the low-rank loss surface is smooth and that Fast Forwarding along a given direction reduces the role of that direction later in training.  We also provide evidence that Fast Forward could be an even more versatile tool: although we cannot Fast Forward full-rank standard finetuning, this limitation is not caused solely by decreased effectiveness on non-attention parameters or in higher dimensional subspaces.

\begin{figure*}
\centering
 \begin{subfigure}{0.49\textwidth}
     \includegraphics[width=\linewidth]{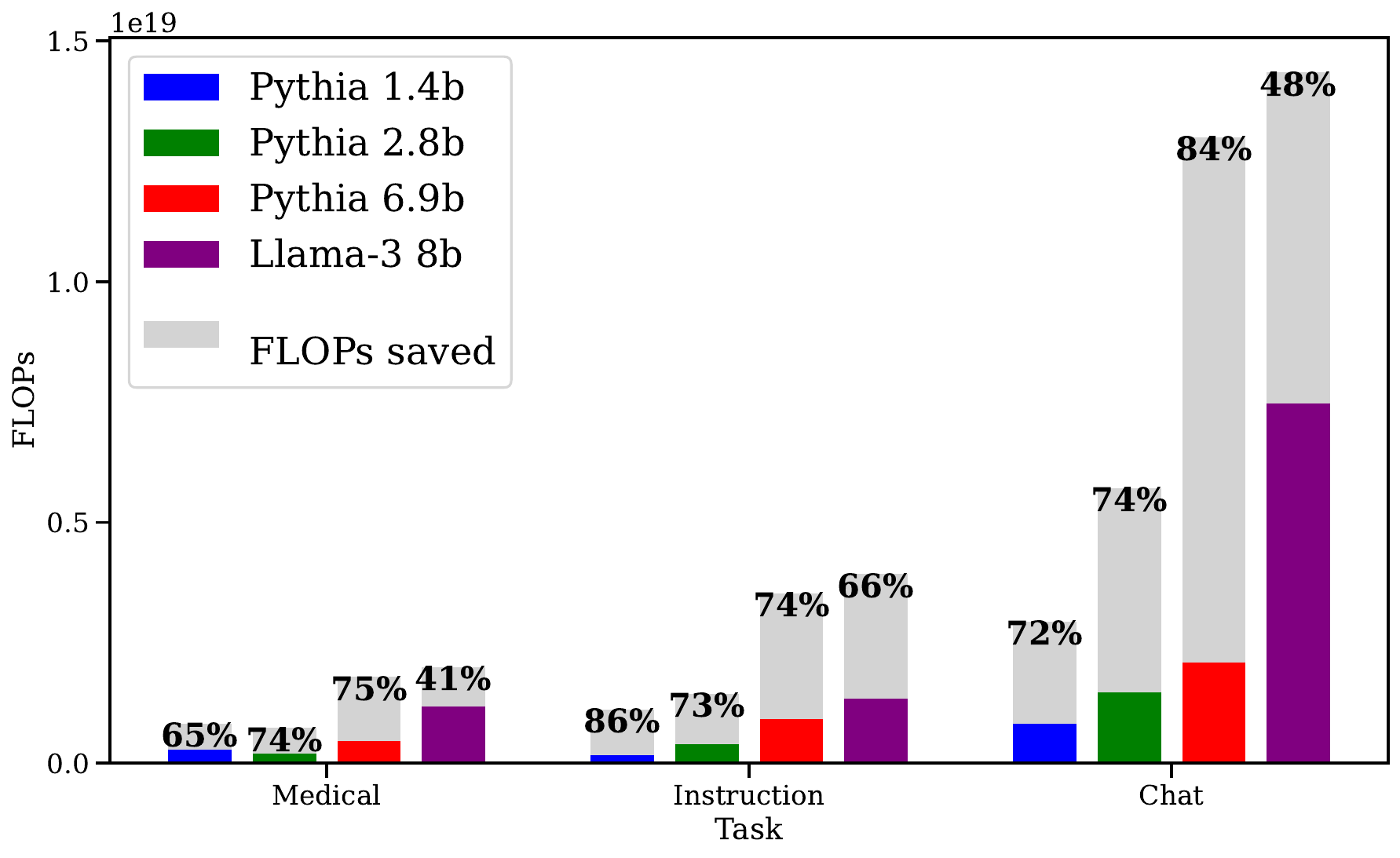}
     \caption{LoRA finetuning}
     \label{fig:lora_results}
 \end{subfigure}\hfill 
  \begin{subfigure}{0.49\textwidth}
     \includegraphics[width=\linewidth]{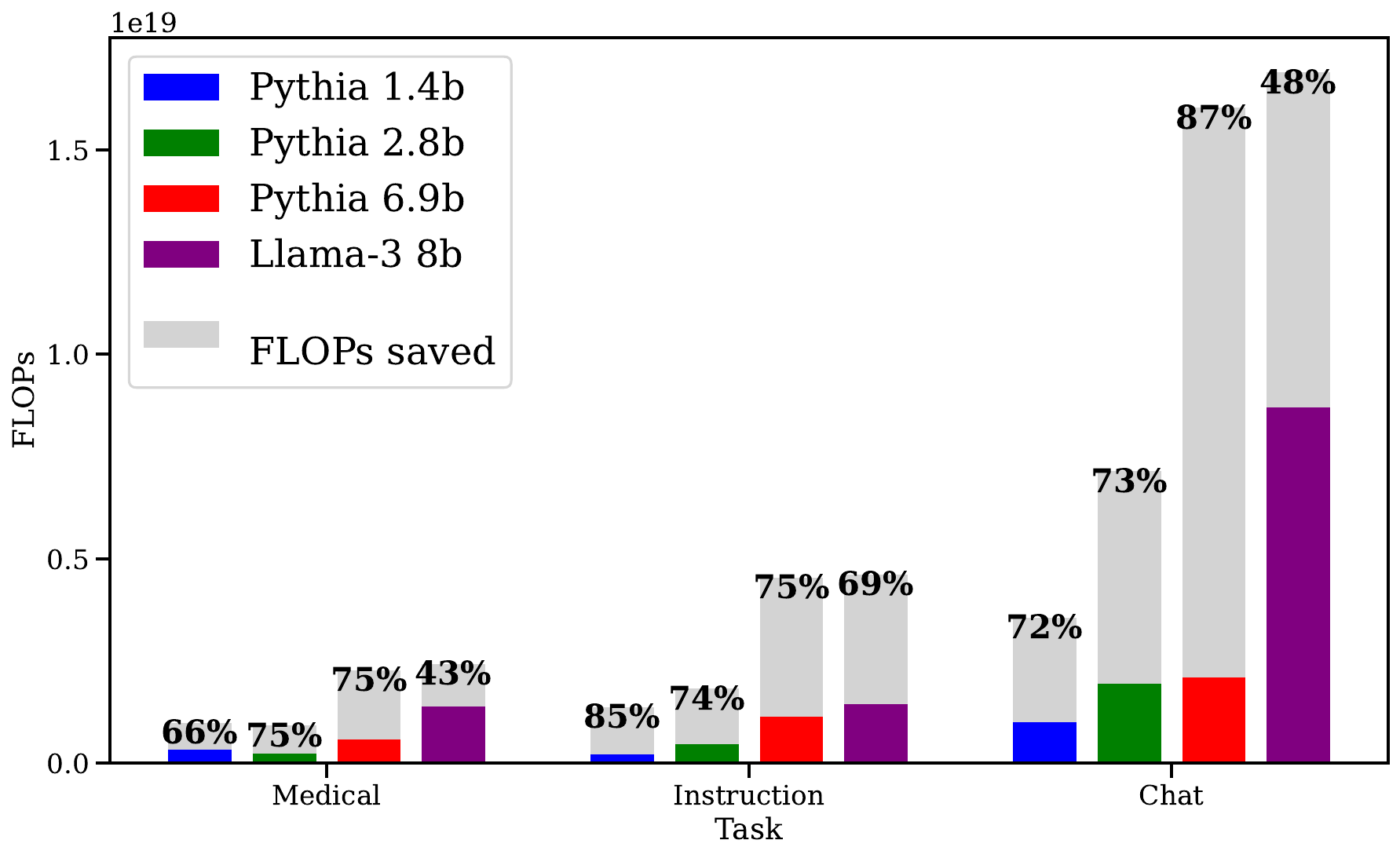}
     \caption{DoRA finetuning}
     \label{fig:dora_results}
 \end{subfigure}
    \caption{The percentage of FLOPs saved during (\protect\subref{fig:lora_results}) LoRA and (\protect\subref{fig:dora_results}) DoRA finetuning with Fast Forward to match test loss after 5 epochs of regular Adam SGD training. Fast Forward saves 41--87\% FLOPs, depending on the task.}
    \label{fig:main_results}
\end{figure*}

\begin{figure}
    \centering
    \includegraphics[width=1\linewidth]{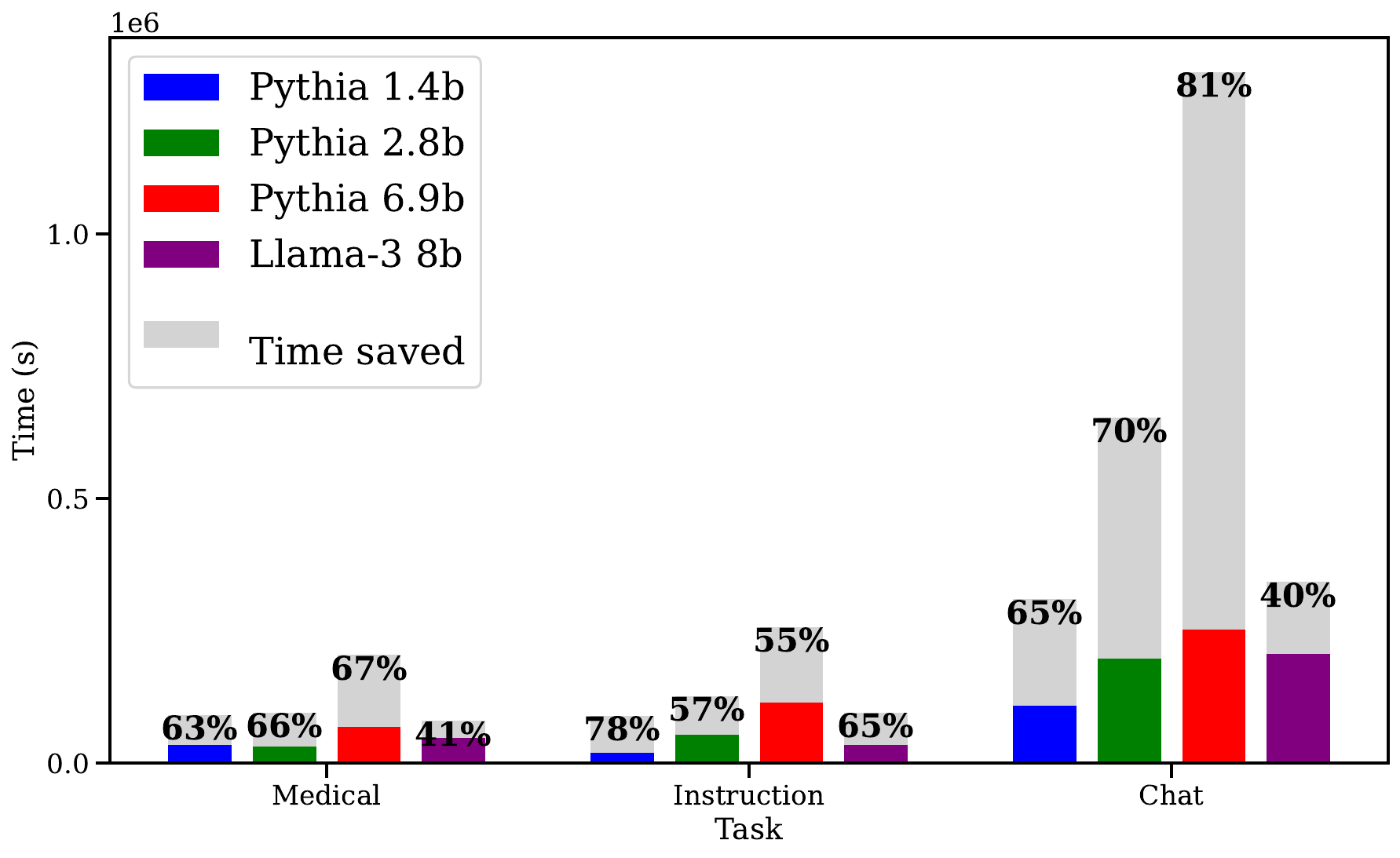}
    \caption{The percentage of train time saved during LoRA finetuning with Fast Forward to match test loss after 5 epochs of regular Adam SGD training. Fast Forward saves 40--81\% of the training time, depending on the task.}
    \label{fig:lora_time}
\end{figure}

\section{Background: Low rank adaptors}
\label{sec:background}

Low rank adaptation (LoRA) \cite{hu2021lora} is a parameter-efficient finetuning method that freezes the LM weights and injects trainable low-rank decompositions into each updated matrix, reducing the number of trainable parameters.
Given a pre-trained parameter setting $\mathbf{W}_0 \in \mathbb{R}^{d\times k}$, LoRA updates the weight with a low-rank decomposition
\begin{equation}
\mathbf{W} = \mathbf{W}_0 + \mathbf{B}\mathbf{A},
\end{equation}
where $\mathbf{B} \in \mathbb{R}^{d\times r}$, $\mathbf{A} \in \mathbb{R}^{r\times k}$ with the low-rank dimension $r \ll d,k$.
Following \citet{hu2021lora}, LoRA only updates the attention matrices.

Weight Decomposed Low-Rank Adaptation (DoRA) \cite{liu2024dora} decomposes the pretrained weight matrix into magnitude and direction, then updates the direction matrix using LoRA.

\section{Fast Forward}
\label{sec:fast_forward}

Our proposal, Fast Forward, is a procedure for accelerating training at regular intervals by selecting an optimal step size with a line search. Following warmup, we apply Fast Forward every $T_{\textrm{interval}} = 6$ optimizer steps, as seen in Figure \ref{fig:method}. During a Fast Forward stage, for each trainable parameter, the difference between weights in the current and previous timesteps is calculated:
\begin{equation} 
\Delta_{\mathbf{W}} = \mathbf{W}_{t} - \mathbf{W}_{t-1} 
\end{equation}
The direction $\Delta_{\mathbf{W}}$ is used to iteratively update $\mathbf{W}_t$. In the $\tau$-th Fast Forward step, the updated weight matrix is given by $\mathbf{W}_t + \tau \Delta_{\mathbf{W}}$. The recursive updates continue until the model's loss on a small validation set $L(X_{\textrm{val}})$ stops improving. When a Fast Forward step causes this validation loss to increase, the Fast Forward stage concludes, and regular optimization resumes for the next $T_{\textrm{interval}}$ steps before reapplying Fast Forward. 


\subsection{Review of related approaches} 

Intermittent schedulers are common in optimization, from classic approaches like the Alternating Direction Method of Multipliers \citep{boyd2011distributed} to modern cyclic hyperparameters. In contrast with our approach of repeatedly maximizing the learning rate, \citet{lialin2023relora} improve low-rank training by repeatedly dropping the learning rate. 

Our work is not the first to train a neural network using line search \citep{vaswani2019painless,truong2018backtracking} or its dual form of trust region optimization \citep{sherborne2023tram}. 
Coordinate descent \citep{wright2015coordinate}, which identifies a coordinate system for the surface and line searches repeatedly along each coordinate, is similar to our approach but does not apply Adam SGD intervals between searches. By retaining historical gradients from the prior SGD stage and changing optimizers when necessary, we can escape saddle points that would trap coordinate descent.

\section{Experiments}
\label{sec:experiments}

\paragraph{Models and data.}

We experiment on three finetuning tasks. For each task, we hold out 1K samples as test and 32 examples as the tiny validation set that determines when to stop Fast Forward. 

\begin{itemize}
    \item \textbf{Medical-domain Tuning:} We train on 37K examples from the Clinical Guidelines corpus \cite{chen2023meditron70b}.
    \item \textbf{Instruction Tuning:} We train on 109K examples from the \href{https://huggingface.co/datasets/ise-uiuc/Magicoder-Evol-Instruct-110K}{decontaminated Evol dataset} \cite{luo2023wizardcoder}
    comprising pairs of code instructions and corresponding outputs.  
    \item \textbf{Chat Tuning:} We train on 208K examples from the \href{https://huggingface.co/datasets/HuggingFaceH4/ultrachat_200k}{filtered ultrachat dataset} \cite{ding2023enhancing}
    of dialogues generated by ChatGPT on various conversational topics.
\end{itemize}


Our models include the open-source Llama-3 8B model \cite{llama3modelcard} as well as the  1.4B, 2.8B, and 6.9B parameter  models from the Pythia suite \cite{biderman2023pythia}, a family of autoregressive transformer language models trained on the Pile dataset \cite{pile}. We train these models using the next token prediction objective for each finetuning corpus. For the instruction tuning task, loss is only based on response completion. Further training hyper-parameters are specified in Appendix \ref{app:parameters}.

\paragraph{Training and Evaluation Procedure}

For a given model and dataset pair, we finetune the model using standard low-rank training for 5 epochs as a baseline, recording the final loss $L_{\mathbf{W}_f}(X_{\textrm{test}})$ as a target and the total training time and number of floating point operations (FLOPs) performed during training. We assume a 1:2 ratio of FLOPs between forward and backward passes \cite{kaplan2020scaling, hoffmann2022training}.



We then retrain the model with intermittent Fast Forward steps until it reaches a test loss within $\epsilon=10^{-4}$ of $L_{\mathbf{W}_f}(X_{\textrm{test}})$. During this stage, we record the total training time and number of FLOPs from all computation, including Adam SGD updates, inference on the small validation set during Fast Forward, and setting model parameters.




\begin{figure}[!ht]
\centering
     \includegraphics[width=\linewidth]{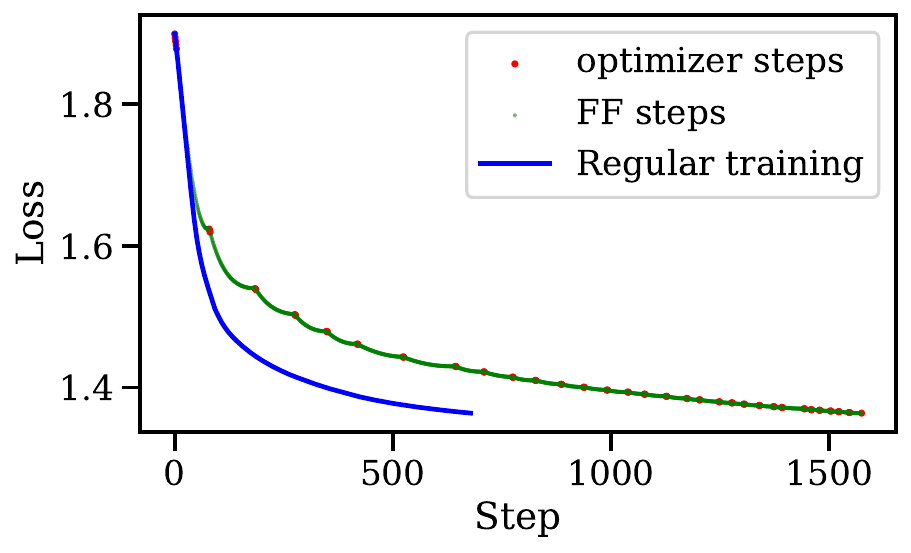}
 \caption{Training Pythia-6.9B on the chat tuning task, with other models in Appx \ref{app:all-models}. Red dots represent SGD steps and green dots represent Fast Forward steps. The blue line shows vanilla Adam SGD training. }
 \label{fig:models_full_training}
\end{figure}



\section{Results}
\label{sec:results}

\textit{Fast Forward accelerates training across low rank methods, in all three datasets and all four models.} 
As Figure \ref{fig:lora_results} shows, 
Fast Forwarding LoRA cuts 41--66\% of finetuning FLOPs for our largest model (Lllama-3 8B) and 65--86\% for our smallest (Pythia 1.4B). Fast Forwarding DoRA, meanwhile, cuts 42--69\% of finetuning FLOPs for Lllama-3 8B and 66--85\% for Pythia 1.4B (Figure \ref{fig:dora_results}). In terms of train time, Fast Forwarding LoRA cuts 63--78\% of Pythia 1.4b training time and 41--65\% of Llama-3 8b training time (Figure \ref{fig:lora_time}). Although Fast Forward is more effective on Pythia than Llama, we see no clear trend as to whether it is generally more effective at smaller scales. 

As seen in Figure~\ref{fig:models_full_training}, Fast Forward accelerates segments of training by simulating predicted SGD steps. Although it requires more total steps than vanilla training (counting SGD interval steps and simulated Fast Forward steps), the efficiency of Fast Forward leads to substantial reductions in computation cost---and Appendix \ref{sec:how-soon} suggests that we could see even greater gains from Fast Forwarding even more often. 
Fast Forward is more effective earlier in training (see Appendix \ref{sec:how_long} for details), but below we find that even after training converges, we see substantial savings.


\subsection{FF does not harm long-term accuracy.}
\label{sec:results_longterm}

Many efficiency methods accelerate training until a fixed threshold accuracy, but ultimately harm final performance. We find that Fast Forward has no such disadvantage. To check, we finetune the Pythia 1.4B model on the medical domain dataset until the loss stopped improving on the test set. We permanently switch to standard Adam SGD after Fast Forward fails to improve the loss on the 32-sample tiny validation set $L(X_{\textrm{val}})   $ three times in a row---though at this point, training ends after only 6 more SGD steps. Fast Forward converges to a slightly better loss while allowing us to save 56\% of FLOPs.

\subsection{FF does not harm performance on a standard benchmark.}
\label{sec:standard_benchmark}
We evaluated whether FF-trained models match the performance of regularly-trained models on a standard benchmark. On the PubMedQA dataset, we evaluated two Llama-3 8b models that were fine-tuned on the medical domain, one that was regularly trained and one with FF. We tested the models on a subset of 1K examples using few-shot in-context learning. As context, all models received the same prompt of 3 randomly chosen examples, one with the answer yes, one with the answer no and one with the answer maybe, in an arbitrary order. The regularly trained model achieved an accuracy of 49.75\%, whereas FF trained model achieved an accuracy of 50.95\%. This shows that FF does not harm standard benchmark results.







\section{When does Fast Forward work?}
\label{sec:results_fullrank}

Our proposed method alternates between conventional Adam SGD and line search, a classic optimization technique. This approach is not only effective, but exceptionally straightforward---so why is it, to our knowledge, undiscovered and unused in modern neural networks? Both line search and intermittent optimizer schedules are well-understood. Why aren't similar approaches standard practice?

The answer may lie in the recent rise of low-rank methods such as LoRA: \textit{Fast Forward is of little use in full-rank standard training}. Even one simulated step increases loss, wasting any compute used for inference on the validation set. The remainder of this section focuses on our efforts to understand why our method only works with low rank training.

\begin{figure}[t]
 \centering
     \includegraphics[width=0.9\linewidth]{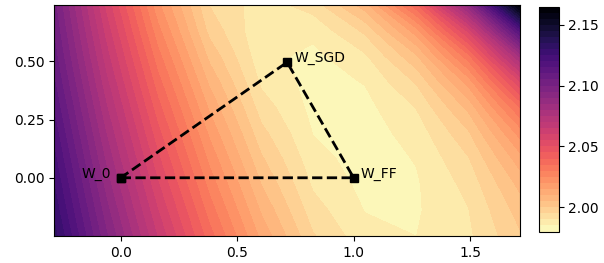}
 \caption{Test loss on the plane intersecting the pretrained model $\mathbf{W}_0$ and the models trained with Adam SGD $\mathbf{W}_{\textrm{SGD}}$, and with Fast Forward $\mathbf{W}_{\textrm{FF}}$. Axis scale corresponds to the norm of differences $\|\mathbf{W}_{\textrm{FF}} -  \mathbf{W}_0\|_2$.
 }
 \label{fig:loss_surface}
 \vspace{-4mm}
\end{figure}

\subsection{Why can't we Fast Forward at full rank?}
\label{sec:results_full_rank_why}


Fast Forward takes advantage of a relatively simple loss surface structure that does not rely on nonlinear paths around barriers. The LoRA loss surface shown in Figure \ref{fig:loss_surface} is roughly convex on the plane that intersects both Fast Forward and SGD directions. Although SGD travels a similar distance, Fast Forward finds a flatter point central to its basin.

\begin{figure}[t]
    \centering
    \includegraphics[width=0.85\linewidth]
    {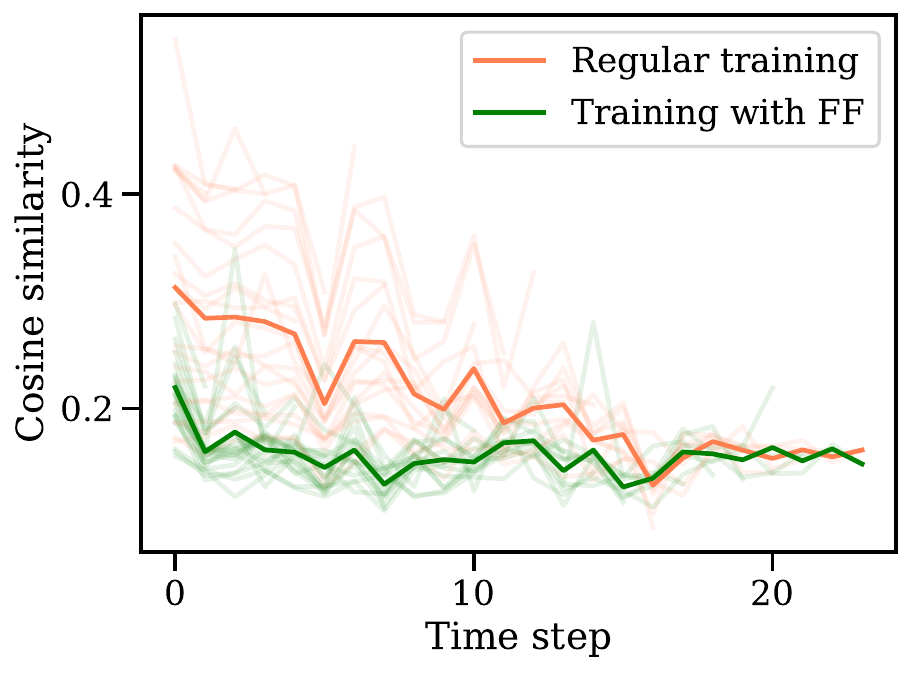}
    \vspace{-3mm}
    \caption{The cosine similarity between gradients in different time steps, in regular training and training with Fast Forward. At each timestep, we measure similarity between the current gradient and all previous saved gradients (shown individually in transparent lines). Fast Forward leads to lower average similarity (shown in opaque lines) with previous gradients.}
    \label{fig:gradients_orthogonlaity}
\end{figure}

Given these conditions, which limit interactions between bases, Fast Forward accelerates to a \textit{persistently} good value for some direction on the surface. As shown in Figure~\ref{fig:gradients_orthogonlaity}, once we Fast Forward in a particular direction, subsequent optimizer steps become less similar to previous steps; because Fast Forward accelerates learning for a specific direction, future optimization steps no longer need to search in that direction.

Attempting to understand the conditions that permit Fast Forward to speculatively optimize along a direction, we consider---and subsequently reject---two hypotheses explaining its failure under standard training. Recall the two primary differences between standard training and LoRA or DoRA: the low-dimensional subspace and the restriction to training only attention matrices. First, we posit that Fast Forward functions only at low rank, meaning that its gains deteriorate as we increase rank. Second, we posit that restricting training to attention will permit Fast Forward even at full rank. As we will demonstrate in this section, neither prediction holds in practice, forcing us to conclude that Fast Forward ceases to improve performance due to the projection matrices added by LoRA.

\begin{figure}[t]
    \centering
    \includegraphics[width=1\linewidth]{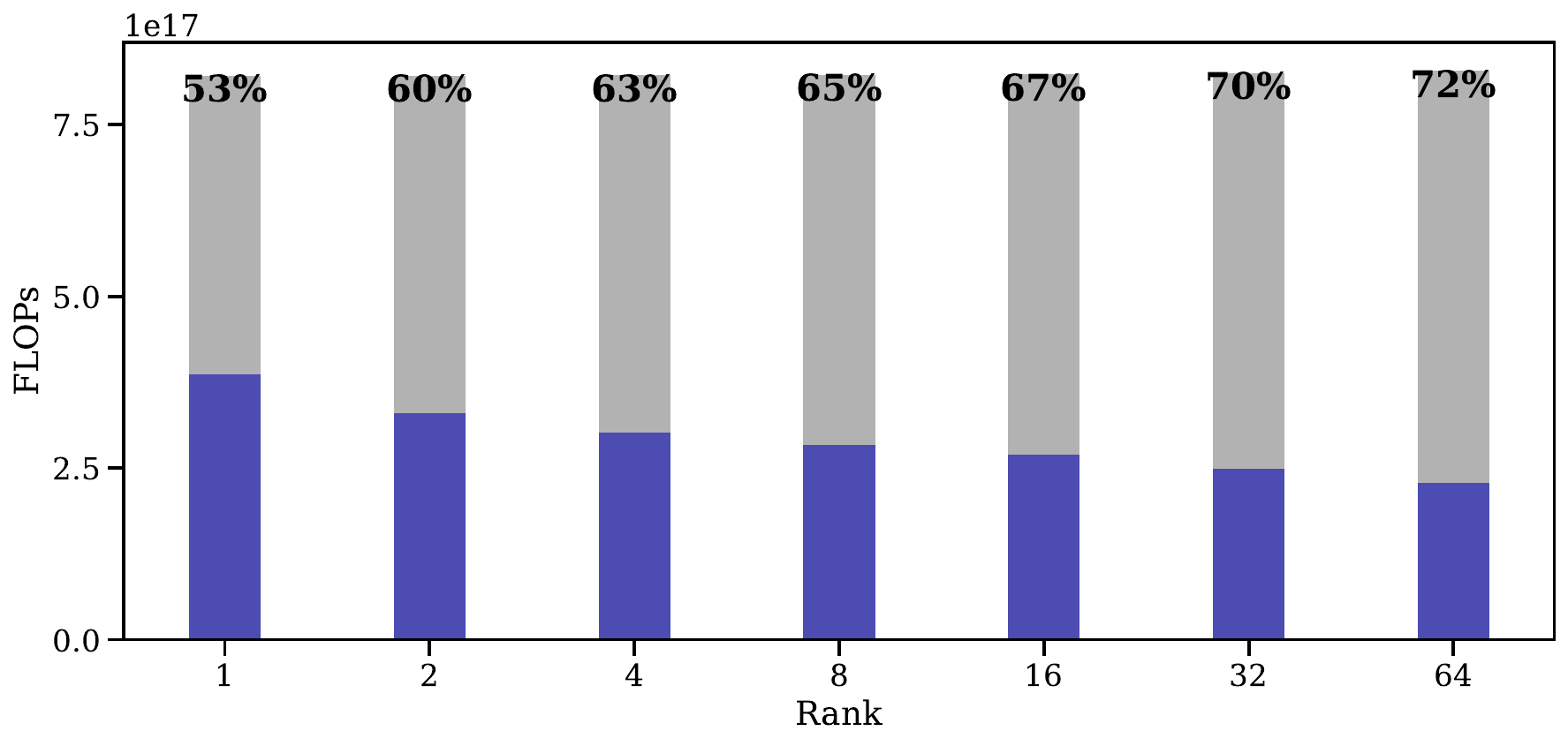}
    \caption{The total number of FLOPs consumed during training Pythia 1.4B on the clinical finetuning task for different LoRA ranks. Gray area is the compute saved with Fast Forward.}
    \label{fig:different_lora_ranks}
\end{figure}

\begin{figure}[t]
    \centering
    \includegraphics[width=1\linewidth]{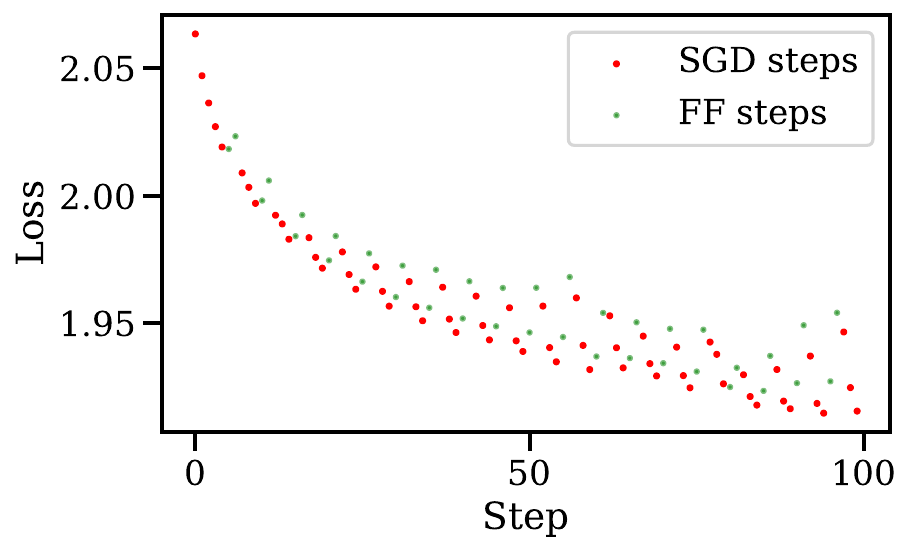}
    \caption{Test loss for full-rank standard finetuning restricted to attention layers. Each time we Fast Forward, loss increases immediately at the first simulated step.}
    \label{fig:ff_attention_learning}
\end{figure}

\paragraph{Fast Forward works better as we increase the rank of LoRA. } Because Fast Forward fails on full-rank standard training, we might assume that it generally degrades with increased rank. In other words, as we add dimensions to LoRA, Fast Forward would become less effective until it stops working---explaining its failure in full-rank standard training. To the contrary, Figure \ref{fig:different_lora_ranks} illustrates that the efficiency gains from Fast Forward increase monotonically with rank between 1 and 64 dimensions. In addition, we compared regular training to FF training on LoRA full rank setting (we set the LoRA rank $r$ to be equal to the weight matrix $\mathbf{W}$ dimension). On the Pythia 410m model, FF training reduced 74\% of the FLOPs compared to regular training.
 Therefore, we cannot confirm the hypothesis that Fast Forward fails at full rank training because of the dimension.

\paragraph{Fast Forward doesn't work in full rank settings even when restricted to the attention layer.} Can we be sure that Fast Forward requires the low-dimensional subspace of LoRA? LoRA doesn't only reduce the dimension of training, but also restricts weight movements to the attention layers. We therefore consider, but ultimately reject, the hypothesis that Fast Forward takes advantage of this module constraint rather than low dimensionality. As seen in Figure~\ref{fig:ff_attention_learning}, Fast Forward performs  poorly for full-rank standard finetuning \textit{even when restricting updates to the attention matrices}.


\section{Conclusions and Future Work}
\label{sec:future}

We have presented Fast Forward, a simple approach to accelerating low-rank finetuning. Our method reduces cost by 41--87\% FLOPs in matching 5-epoch Adam SGD loss and 56\% at loss convergence. 

A variety of tweaks on our approach are likely to convey further benefits. Rather than using a fixed tiny validation set throughout training, we might introduce new ways of sampling that introduce little overhead but avoid the possibility of overfitting. Other future work may schedule the SGD interval lengths dynamically or predict the optimal duration for Fast Forward, reducing the required number of inference runs on the validation set.

One reading of our findings is that current optimizers like Adam are poorly designed for low-rank approaches. Future optimizers might improve these standard approaches by aligning momentum with the known low-rank basis or applying other methods that select better step sizes at low dimensions.
As one example of an issue with momentum alignment, \citet{hayou_lora_2024} proves that LoRA's uniform learning rate is suboptimal and proposes a fix; however, their modified optimizer requires substantial tuning and may not significantly increase training efficiency for a full training run.


\section*{Limitations}

Our approach, as discussed, does not appear to accelerate full-rank standard training and therefore may not be usable as-is when training from scratch.  Although our experiments are limited to finetuning, low-rank pretraining methods like GaLoRe \citep{zhao2024galore}  might also benefit from this type of acceleration. 

Fast Forward is highly efficient, but the acceleration step itself is serialized. In order to improve its parallelization, further work is needed. While we do not focus on this efficiency improvement, other optimizers that search different subspaces at regular intervals have been parallelized \citep{wei2012distributed}, and perhaps Fast Forward could be as well.

To measure compute, we use FLOPs, a metric that does not always reflect caching and other overhead elements and does not take into account parallelization. 

All experiments are conducted on English language datasets with conventional autoregressive LLMs. Our results are also limited to next token prediction finetuning objectives. 


\section*{Acknowledgements}

This work was enabled in part by a gift from the Chan Zuckerberg Initiative Foundation to establish the Kempner Institute for the Study of Natural and Artificial Intelligence.

We thank Nikhil Vyas and Sham Kakade for invaluable discussion that shaped this work.




\bibliography{custom}
\newpage
\appendix

\section{Loss throughout training on all models} \label{app:all-models}

\begin{figure}[!ht]
 \begin{subfigure}{0.49\textwidth}
     \includegraphics[width=\linewidth]{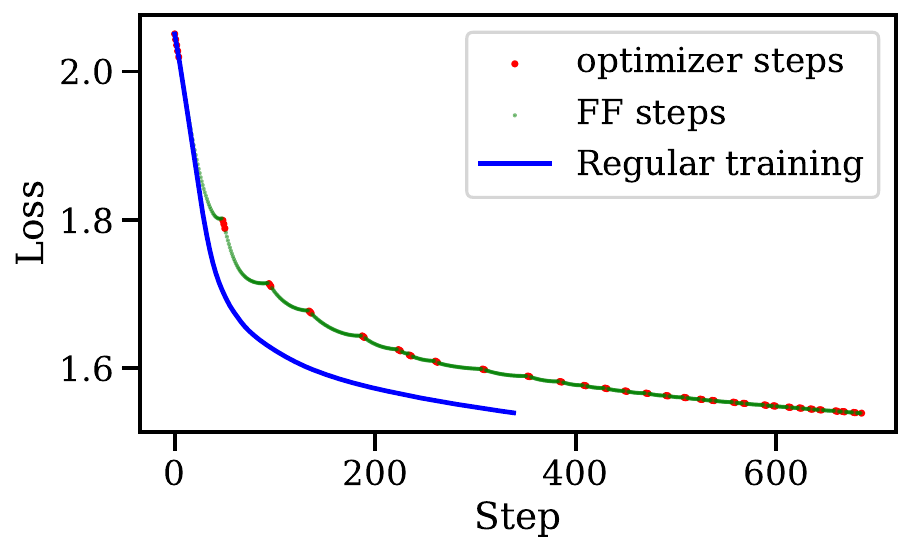}
     \caption{Pythia 1.4B}
     \label{fig:1_4b_full_training}
 \end{subfigure}
 \hfill
 \begin{subfigure}{0.49\textwidth}
     \includegraphics[width=\linewidth]{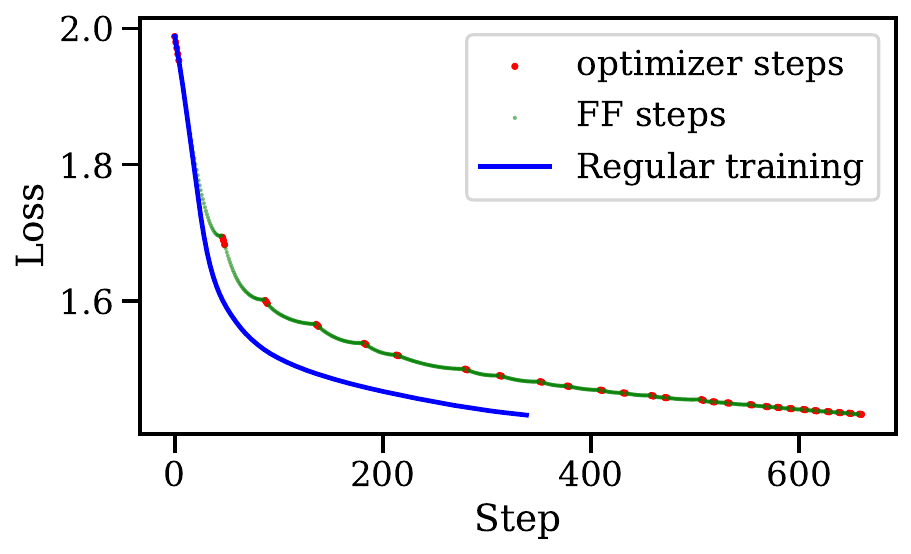}
     \caption{Pythia 2.8B}
     \label{fig:2_8b_full_training}
 \end{subfigure}
 \begin{subfigure}{0.49\textwidth}
     \includegraphics[width=\linewidth]{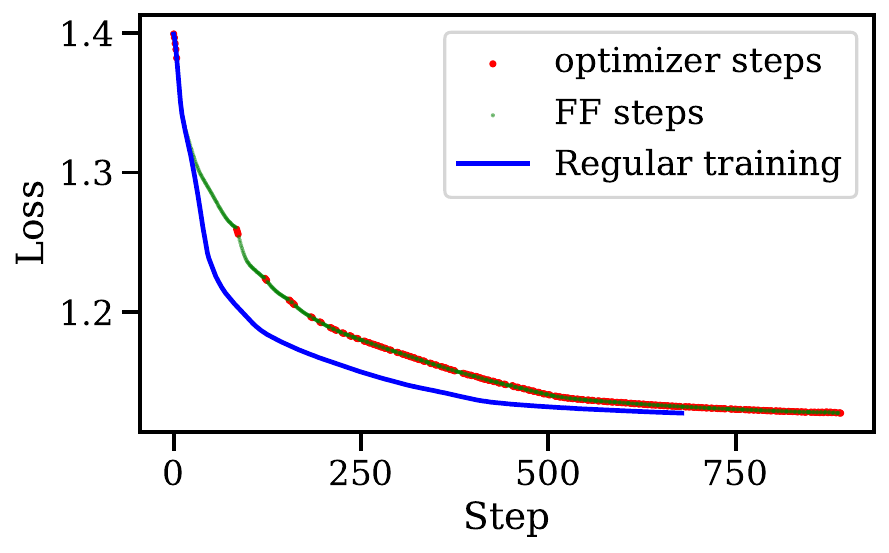}
     \caption{LLama-3.8B}
     \label{fig:llama_full_training}
 \end{subfigure}
 \caption{Training each model on the chat tuning task. Red dots represent SGD steps and green dots represent Fast Forward steps. The blue line shows vanilla Adam SGD training. }
 \label{fig:models_full_training_appx}
\end{figure}

Figure \ref{fig:models_full_training_appx} shows that all models exhibit a similar pattern wherein the simulated steps briefly accelerate the drop in loss between each SGD interval.

\section{How long can we Fast Forward?}
\label{sec:how_long}

\begin{figure}[t]
    \centering
    \includegraphics[width=1\linewidth]{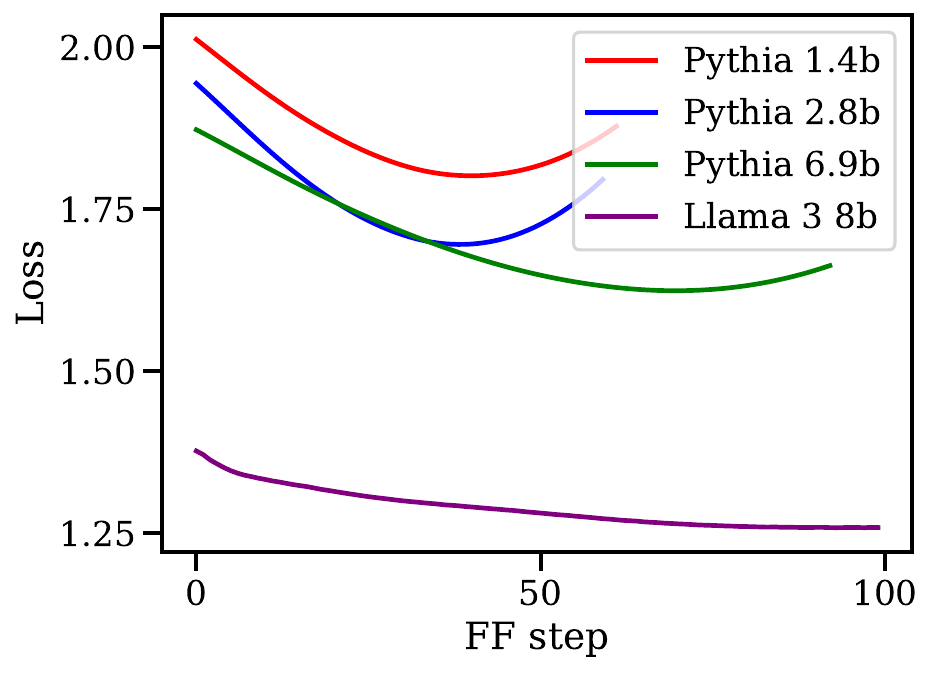}
    \caption{Test loss for the first Fast Forward on the chat tuning task, run for a duration of 100 simulated steps. Within this range, the resulting loss curves are convex with respect to the number of Fast Forward steps.}
    \label{fig:ff_more_steps}
\end{figure}

Figure~\ref{fig:ff_more_steps} illustrates that the loss is convex under Fast Forward, meaning that we can identify a vertex by searching linearly until loss begins to rise on our 32-sample tiny validation set.
The vertex along a particular update direction is that direction's locally optimal step size---that is, the step size that leads to the greatest immediate decrease in loss. 

\begin{figure}[t]
    \centering
    \includegraphics[width=1.0\linewidth]{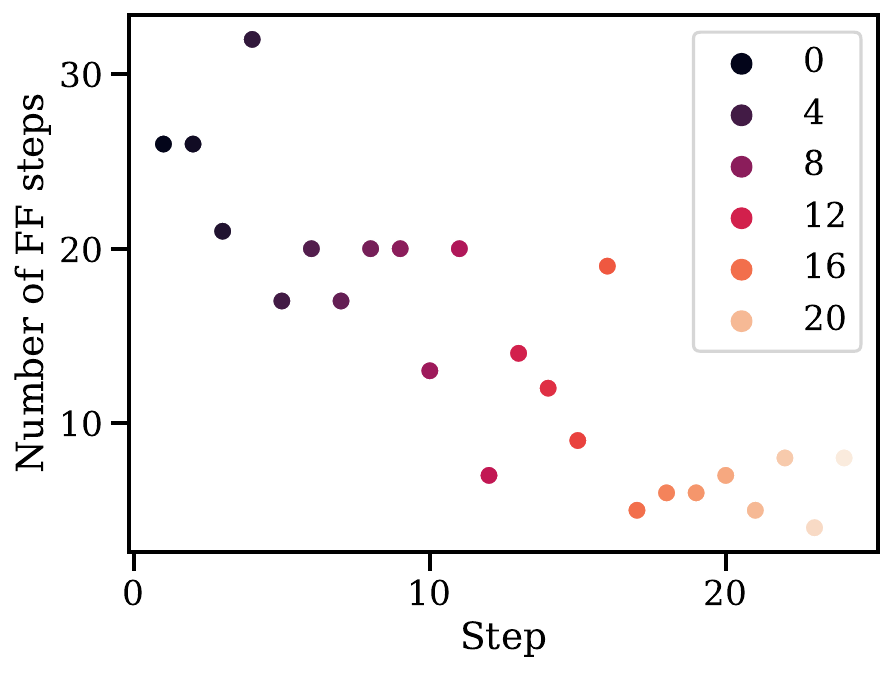}
    \caption{The optimal number of Fast Forward steps performed as a function of the Fast Forward stage during training. Darker points represent stages earlier in training. As training continues, the number of Fast Forward steps performed decreases.}
    \label{fig:ff_steps_ff_phase}
\end{figure}

At each step during normal training, the gradient---and consequently the change in weights $\Delta_\mathbf{W}$ between steps---is modified by Adam and other optimizer components.  We consider a number of possible factors determining the optimal number of simulated steps before the loss begins to increase, $\tau^* = \arg\max_{\tau} \mathbf{W}_t + \tau \Delta_\mathbf{W}$. As seen in Figure \ref{fig:ff_steps_ff_phase}, $\tau^*$ declines over the course of training, meaning that Fast Forward becomes less productive.

\begin{figure}[h]
    \centering
     \begin{subfigure}{0.49\textwidth}
        
    \includegraphics[width=1\linewidth]{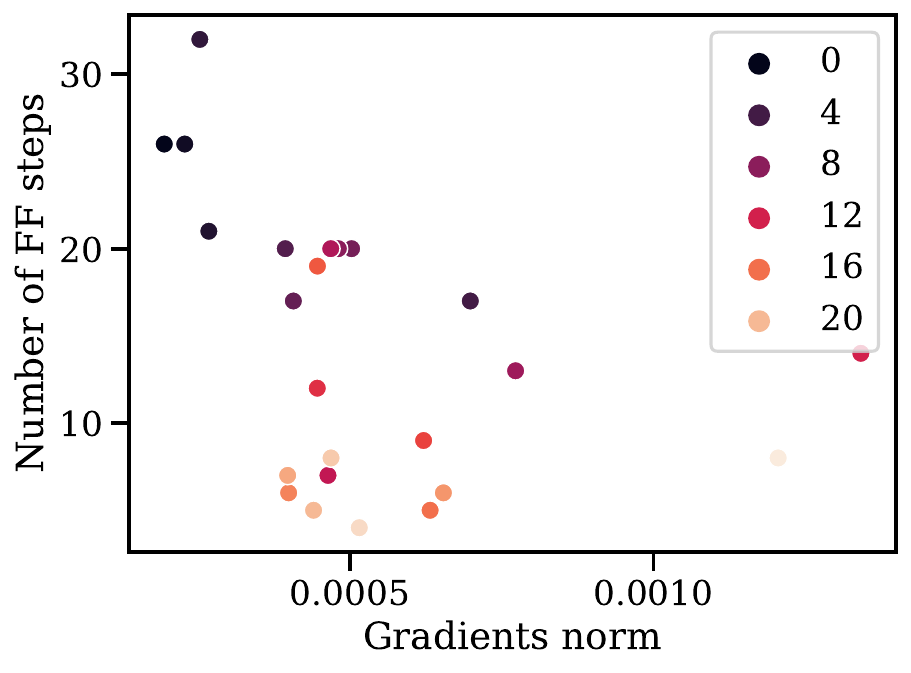}
    \caption{The number of Fast Forward steps performed as a function of the gradients' norm. The norms are of the gradients right before a new Fast Forward stage is performed. }
    \label{fig:ff_steps_as_grad_norm}
    \end{subfigure}
    
    \medskip
    
 \begin{subfigure}{0.49\textwidth}
    \centering
    \includegraphics[width=1\linewidth]{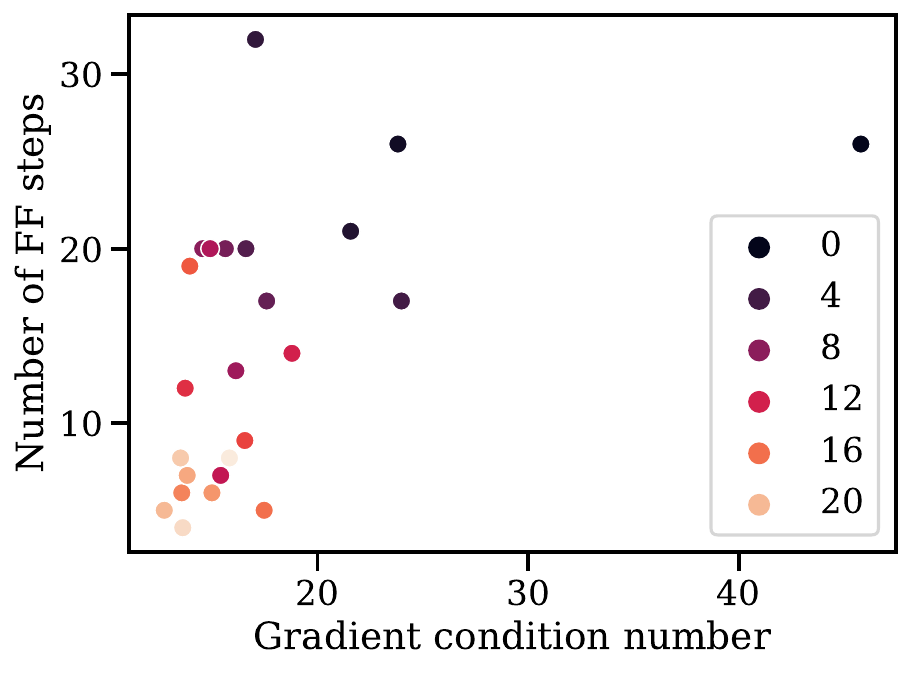}
    \caption{The number of Fast Forward steps performed as a function of the gradients' condition number. The condition numbers are of the gradients right before a new Fast Forward stage is performed. }
    \label{fig:ff_steps_condition}
\end{subfigure}
\caption{Potential factors in determining the optimal Fast Forward step count. Darker points represent stages earlier in training, whereas lighter points represent stages later in training. The experiment was performed on the Pythia 1.4B model on medical finetuning task.}
\label{fig:ff_steps}
\end{figure}

Although we consider some possible factors determining the optimal Fast Forward duration in Figures \ref{fig:ff_steps_as_grad_norm} and \ref{fig:ff_steps_condition}, neither the norm nor the condition number of the gradient provide predictive power beyond knowing the current training timestep. While both are clearly correlated, that correlation depends on a confounder with both factors: the duration of training.


\section{Consistency of gradients}

Based on the conceptualization of Fast Forward as literally being a simulation of the next steps of training, one might expect that it would work better if the gradients were similar for different batches. Under this expectation, the direction that encompasses multiple batches will generalize across the entire dataset and therefore work better than a direction that only applies to the most recent batch selected. Such a heuristic could choose when to Fast Forward by identifying times when we could execute more steps.

The fundamental assumption here is that directions on the loss surface which are ``wide'' (that is, applying even under slight distribution changes, as occur between batches) must also be ``long'' (that is, we can follow that gradient further and continue to improve loss).

\begin{figure}
    \centering
    \includegraphics[width=1\linewidth]{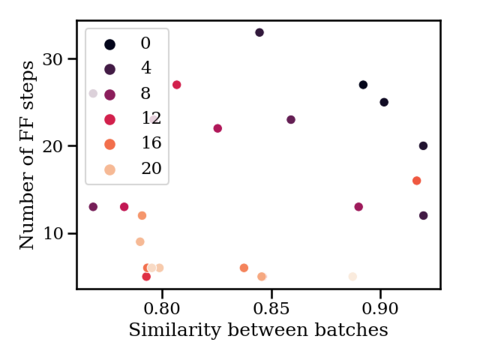}
    \caption{Gradient consistency across batches vs the length of the optimal Fast Forward stage. Gradient consistency is given by cosine similarity between the gradients of different batches, as measured immediately before a Fast Forward Stage.}
    \label{fig:gradient-consistency}
\end{figure}

To the contrary, Figure \ref{fig:gradient-consistency} shows no significant correlation between batch-wise gradient consistency and the optimal length of a Fast Forward. The implications around loss surface geometry are intriguing: Even the most broadly applicable gradient steps might be useful only briefly, and immediately encounter obstacles that require nonlinear paths. 

\section{How soon can we Fast Forward?}
\label{sec:how-soon}

How long should we train for in between Fast Forwards? Here, we identify the point in training at which a conventional optimizer has temporarily settled into a consistent direction which can be extrapolated without damaging performance. The more Fast Forward steps we can take, the more effective Fast Forward is at a given point in time. 
We set different values for $T_\textrm{interval}$, from 1 to 10, and measured the number of Fast Forward steps performed in that point of training. Experiment was performed on the Pythia 1.4B model on the medical finetuning task.

\begin{figure}[t]
    \centering
    \includegraphics[width=1.0\linewidth]{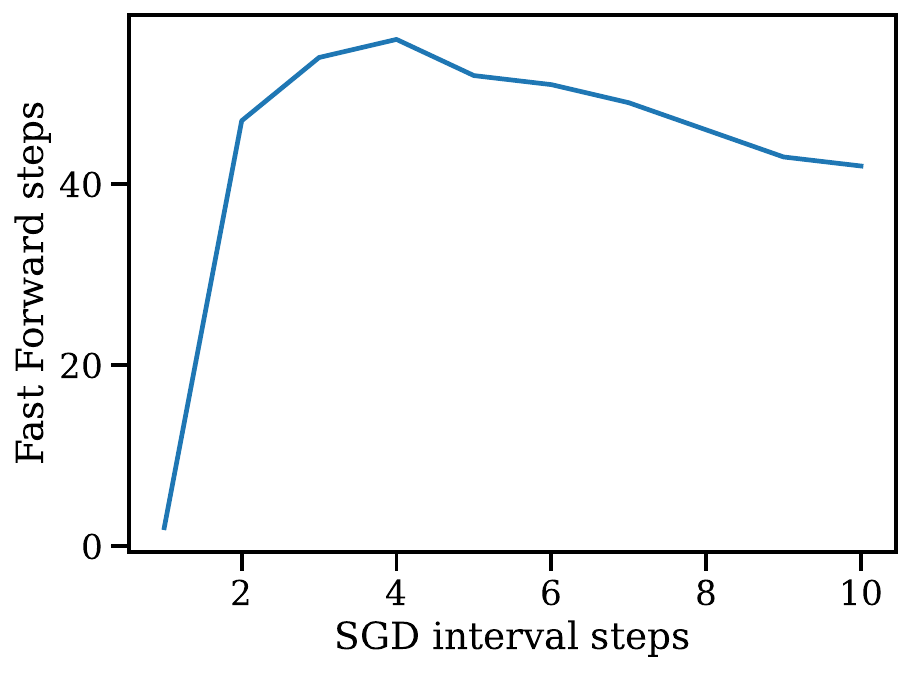}
    \caption{Optimal number of Fast Forward steps performed at the second Fast Forward stage, as a function of the number of SGD steps performed in the interval since the last Fast Forward stage. (One step is equivalent to extending the previous FF stage.)}
    \label{fig:FFvsOptim}
\end{figure}

Figure \ref{fig:FFvsOptim} illustrates the relationship between the duration of the SGD interval stage and the duration of the subsequent Fast Forward stage. Before the second Fast Forward stage, training for an interval of up to 4 SGD steps extends the optimal number of Fast Forward steps. Continuing to run the default optimizer further begins to limit Fast Forward.

Note that we can start benefiting from Fast Forward---that is, loss decreases a nonzero amount---immediately after running a pair of SGD interval steps. We might therefore save even more compute, depending on the setting, by running SGD for even less time early in training. 

\section{Training hyper parameters}
\label{app:parameters}
Hyperparameters used for training different models and tasks. Table \ref{tab:med_param} shows the hyper-parameters used for medical-domain tuning, Table \ref{tab:instruct_param} shows the hyper-parameters used for instruction tuning and Table \ref{tab:chat_param} shows the hyper-parameters used for chat tuning.
\begin{table}[]
\small
\begin{tabular}{@{}ccccc@{}}
\toprule
 &
  \begin{tabular}[c]{@{}c@{}}Learning\\ rate\end{tabular} &
  \begin{tabular}[c]{@{}c@{}}Micro batch\\  size\end{tabular} &
  \begin{tabular}[c]{@{}c@{}}Global batch\\  size\end{tabular} &
  \begin{tabular}[c]{@{}c@{}}LoRA\\  r\end{tabular} \\ \midrule
\begin{tabular}[c]{@{}c@{}}Pythia\\ 1.4b\end{tabular} & 4.0e-5 & 2 & 128 & 8 \\ \midrule
\begin{tabular}[c]{@{}c@{}}Pythia\\ 2.8b\end{tabular} & 4.0e-5 & 1 & 128 & 8 \\ \midrule
\begin{tabular}[c]{@{}c@{}}Pythia\\ 6.9b\end{tabular} & 4.0e-5 & 1 & 128 & 8 \\ \midrule
\begin{tabular}[c]{@{}c@{}}Llama-3\\ 8b\end{tabular}  & 4.0e-5 & 1 & 128 & 8 \\ \bottomrule
\end{tabular}
\caption{Medical tuning hyper-parameters}
\label{tab:med_param}
\end{table}

\begin{table}[t]
\small
\begin{tabular}{@{}ccccc@{}}
\toprule
 &
  \begin{tabular}[c]{@{}c@{}}Learning\\ rate\end{tabular} &
  \begin{tabular}[c]{@{}c@{}}Micro batch\\  size\end{tabular} &
  \begin{tabular}[c]{@{}c@{}}Global batch\\  size\end{tabular} &
  \begin{tabular}[c]{@{}c@{}}LoRA\\  r\end{tabular} \\ \midrule
\begin{tabular}[c]{@{}c@{}}Pythia\\ 1.4b\end{tabular} & 5.0e-6 & 2 & 64 & 8 \\ \midrule
\begin{tabular}[c]{@{}c@{}}Pythia\\ 2.8b\end{tabular} & 5.0e-6 & 1 & 64 & 8 \\ \midrule
\begin{tabular}[c]{@{}c@{}}Pythia\\ 6.9b\end{tabular} & 5.0e-6 & 1 & 64 & 8 \\ \midrule
\begin{tabular}[c]{@{}c@{}}Llama-3\\ 8b\end{tabular}  & 5.0e-6 & 1 & 64 & 8 \\ \bottomrule
\end{tabular}
\caption{Instruction tuning hyper-parameters}
\label{tab:instruct_param}
\end{table}

\begin{table}[t]
\small
\begin{tabular}{@{}ccccc@{}}
\toprule
 &
  \begin{tabular}[c]{@{}c@{}}Learning\\ rate\end{tabular} &
  \begin{tabular}[c]{@{}c@{}}Micro batch\\  size\end{tabular} &
  \begin{tabular}[c]{@{}c@{}}Global batch\\  size\end{tabular} &
  \begin{tabular}[c]{@{}c@{}}LoRA\\  r\end{tabular} \\ \midrule
\begin{tabular}[c]{@{}c@{}}Pythia\\ 1.4b\end{tabular} & 2.0e-5 & 2 & 512 & 64 \\ \midrule
\begin{tabular}[c]{@{}c@{}}Pythia\\ 2.8b\end{tabular} & 2.0e-5 & 2 & 512 & 64 \\ \midrule
\begin{tabular}[c]{@{}c@{}}Pythia\\ 6.9b\end{tabular} & 2.0e-5 & 1 & 512 & 64 \\ \midrule
\begin{tabular}[c]{@{}c@{}}Llama-3\\ 8b\end{tabular}  & 2.0e-5 & 1 & 512 & 64 \\ \bottomrule
\end{tabular}
\caption{Chat tuning hyper-parameters}
\label{tab:chat_param}
\end{table}

\end{document}